\documentclass{article}

\usepackage[final]{neurips_2023}

\usepackage[utf8]{inputenc} 
\usepackage[T1]{fontenc}    
\usepackage{hyperref}       
\usepackage{url}            
\usepackage{booktabs}       
\usepackage{amsfonts}       
\usepackage{nicefrac}       
\usepackage{microtype}      
\usepackage{xcolor}         

\usepackage{amsmath}
\usepackage{amsfonts}
\newtheorem{theorem}{Theorem}
\newtheorem{lemma}{Lemma}
\newtheorem{assumption}{Assumption}

\usepackage{multirow}
\usepackage{mathrsfs}
\usepackage{graphics}
\usepackage{subfigure}
\usepackage{wrapfig}
\usepackage{epstopdf}
\usepackage{pdfpages}
\usepackage{diagbox}
\usepackage{ulem}
\usepackage{CJKulem}
\usepackage{MnSymbol}

\usepackage{tabularx}
\usepackage{diagbox}
\usepackage{makecell}

\title{D-Separation for Causal Self-Explanation}

\author{
Wei Liu$^1$ \quad Jun Wang$^2$\footnotemark[1] \quad Haozhao Wang$^1$\thanks{Corresponding authors. This paper is a collaboration between Intelligent and Distributed Computing Laboratory, Huazhong University of Science and Technology and \href{https://www.iwudao.tech/}{iWudao Tech}.} \quad Ruixuan Li$^1$\footnotemark[1]\\ \textbf{Zhiying Deng}$^1$ \quad \textbf{Yuankai Zhang}$^1$ \quad \textbf{Yang Qiu}$^1$
\\$^1$School of Computer Science and Technology, Huazhong University of Science and Technology\\  $^2$iWudao Tech\\
$^1$\texttt{\{idc\_lw, hz\_wang, rxli, dengzhiyingdd, yuankai\_zhang, anders\}@hust.edu.cn } \\ $^2$\texttt{jwang@iwudao.tech}\\
}

\begin{document}

\normalem

\maketitle

\begin{abstract}
Rationalization is a self-explaining framework for NLP models. Conventional work typically uses the maximum mutual information (MMI) criterion to find the rationale that is most indicative of the target label. 
However, this criterion can be influenced by spurious features that correlate with the causal rationale or the target label. Instead of attempting to rectify the issues of the MMI criterion, we propose a novel criterion to uncover the causal rationale, termed the Minimum Conditional Dependence (MCD) criterion, which is grounded on our finding that the non-causal features and the target label are \emph{d-separated} by the causal rationale. By minimizing the dependence between the unselected parts of the input and the target label conditioned on the selected rationale candidate, all the causes of the label are compelled to be selected. In this study, we employ a simple and practical measure of dependence, specifically the KL-divergence, to validate our proposed MCD criterion.  
Empirically, we demonstrate that MCD improves the F1 score by up to $13.7\%$ compared to previous state-of-the-art MMI-based methods.
Our code is available at: \url{https://github.com/jugechengzi/Rationalization-MCD}.
\end{abstract}

\section{Introduction}\label{sec:introduction}

\begin{wrapfigure}[8]{R}{0.7\columnwidth}
\vspace{-30pt}
    \includegraphics[width=0.7\columnwidth]{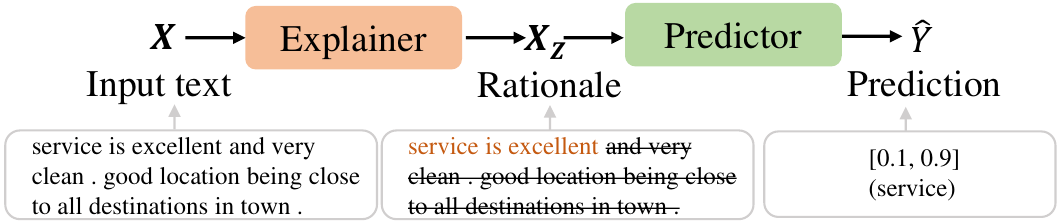}
  \caption{The standard rationalization framework RNP. $X$ is the original full text. $X_Z$ is the selected rationale candidate and $\hat{Y}$ is the predictor's output.}
  \label{fig:rnp}
\end{wrapfigure}
With the success of deep learning, there is growing concern about the interpretability of deep learning models, particularly as they are rapidly being deployed in various critical fields \citep{lipton2016mythos}. Ideally, the explanation for a prediction should be both faithful (reflecting the model's actual behavior) and plausible (aligning with human understanding) \citep{Unirex}.

Post-hoc explanations, which are trained separately from the prediction process, may not faithfully represent an agent's decision, despite appearing plausible \citep{lipton2016mythos}. Sometimes, faithfulness should be considered a prerequisite that precedes plausibility in explanations of neural networks, especially when these networks are employed to assist in critical decision-making processes, as this factor determines the trustworthiness of the explanations. In contrast to post-hoc methods, ante-hoc (or self-explaining) techniques typically offer increased transparency \citep{lipton2016mythos} and faithfulness \citep{interlocking}, as the prediction is made based on the explanation itself.

A model-agnostic ante-hoc explanation framework, called Rationalizing Neural Predictions (RNP), was proposed by \citet{rnp} and is also known as rationalization. RNP utilizes a cooperative game between an explainer and a predictor, where the explainer identifies a human-interpretable subset of the input (referred to as rationale) and passes it to the subsequent predictor for making predictions, as shown in Figure~\ref{fig:rnp}. The explainer and predictor are trained cooperatively to maximize prediction accuracy. A significant advantage of RNP-based rationalization is its certification of exclusion, which guarantees that any unselected part of the input has no contribution to the prediction. This property ensures the maintenance of faithfulness, enabling us to focus solely on plausibility~\citep{interlocking}. 
Notably, although RNP was initially proposed in the field of NLP and and its enhancement schemes have primarily been validated using text data, its framework can also be applied to other domains, e.g., explaining image classification \citep{GDM} and graph neural networks \citep{pgexplainer}. 

Previous rationalization methods generally utilize the maximum mutual information (MMI) criterion to determine the rationale, defined as the subset most indicative of the target label. However, this criterion merely uncovers associations rather than causal relationships between the rationale and the label. Consequently, MMI is easily affected by spurious correlations and the plausibility of chosen rationales might be diminished,  even though the rationales still faithfully report the predictor's behavior \citep{invarant}.
In rationalization, there are two stages from which correlations may arise. 
The first type of correlation originates from the process of dataset generation, and we refer to it as feature correlation. A typical example of feature correlation, as pointed out in LIME \citep{LIME}, is that wolves often appear together with snow. Consequently, whether the background features snow or not can serve as a strong indicator for classifying an image as depicting a wolf.
Another instance of feature correlation is demonstrated in the first row of Table~\ref{tab:examples}. Within a beer review, a favorable taste often correlates with an appealing aroma. Comments regarding the taste can serve as a strong indicator for the smell label. The predictor might inadvertently overfit to such correlations, leading to local optima. Consequently, the suboptimal predictor could mislead the explainer to select these spurious correlations. 
Another type of correlation stems from the rationale (mask) selection stage, and we call it mask correlation.
An example of mask correlation is depicted in the second row of Table~\ref{tab:examples}. Consider a situation where the explainer has implicitly learned the category of $X$, and selects a $``$-$"$ for all negative inputs while excluding it from all positive inputs. In this case, the predictor only needs to determine whether the input rationale includes a $``$-$"$ or not. Even though this phenomenon has not been analyzed from the perspective of spurious correlation, it has been observed and named as \textbf{degeneration} in prior research \citep{rethinking}.

Some methods have been developed to address either feature correlation or degeneration separately. INVRAT \citep{invarant} attempts to tackle feature correlation using invariant risk minimization (IRM) \citep{invariant-risk}. The main idea is to emphasize spurious (non-causal) variations by splitting the dataset into distinct environments. However, IRM-based methods have several limitations. For instance, they require strong prior knowledge about the relationships between non-causal and causal features (e.g., the extra labels of non-causal features) in order to divide the dataset \citep{env-part}. Moreover, IRM-based methods are limited to addressing only a finite set of predetermined non-causal features, neglecting the potential existence of numerous unknown non-causal features. In fact, a recent study \citep{env-part} in the field of IRM has theoretically demonstrated that it is nearly impossible to partition a dataset into different environments to eliminate all non-causal features using IRM. Other challenges, such as the tendency to overfit, difficulty in applying to larger models \citep{irm-overfit,impossible-env}, and the marginal shift risk of the input \citep{marginal-shift-risk}, have also been identified within the realm of IRM. Inter\_RAT \citep{interventional} attempts to eliminate feature correlation through \emph{backdoor adjustment}, intervening directly with the confounders. However, it is extremely hard to measure the confounders since they are usually not observable in the dataset.
As for degeneration, although not explicitly associated with spurious correlation until this study, some efforts have been tried to fix the problem. The common idea is to introduce auxiliary modules that have access the full texts to regularize the original explainer \citep{interlocking} or the predictor \citep{interlocking,liufr}. Regularized by these auxiliary modules, the predictor can somewhat disregard the mask correlation, and degeneration is partially alleviated. 

Although these methods have tried to fix the problems resulted from spurious correlations, they are still MMI-based methods and how the problems come into being is not well explored. In this study, we first identify the two stages that the spurious correlations may come from, and link the important degeneration problem to a more general mask correlation. Then, we identify that the target label $Y$ and all the non-causal features (including the rationale masks) in the input $X$ are \emph{d-separated} by the causal features, meaning that the non-causal features are independent of $Y$ given the causal features. This leads to a new avenue for addressing feature correlation and mask correlation simultaneously: we only need to penalize the dependence between the target label and the unselected features conditioned on the selected rationale candidate, such that all the direct causal features will be included in the selected rationale candidate. Based on this observation, we develop a new criterion for the causal rationale, namely minimum conditional dependence (MCD). Various methods can be adopted to measure the dependence, such as mutual information, the Hilbert-Schmidt Independence Criterion (HSIC) \citep{hsic}, and so on. In this paper, we adopt a simple and practical measurement for independence, the KL-divergence, to verify the effectiveness of the proposed criterion. Then, we conduct experiments on two widely used benchmarks to validate the effectiveness of MCD.

\begin{table*}[t]
\caption{ The examples of feature correlation and mask correlation. Human-annotated rationales are \underline{underlined}. Rationales from RNP are highlighted in \textcolor{red}{red}. } 
\footnotesize
\begin{tabularx}{\linewidth}{X}
	\hline\hline
	\makecell[c]{RNP} \\\hline\hline
	\textbf{Dataset:} Beer-Aroma. \  \textbf{Label:} Positive. \textbf{Predition:} Positive. \textbf{Problem:} feature correlation\\
 \textbf{Text:} the appearance was nice . dark gold with not much of a head but nice lacing when it started to dissipate . \quad 
    \underline{\textcolor{red}{the}  smell was ever so hoppy with a hint of the grapefruit flavor that 's contained within .} \  \textcolor{red}{the taste was interesting , up front tart grapefruit , not sweet in the} least . more like grapefruit rind even . $\cdots \cdots.$
		\\
    \hline
    \hline
    \textbf{Dataset:} Beer-Aroma. \ \textbf{Label:} Negative. \textbf{Predition:} Negative.  \textbf{Problem:} mask correlation\\
 \textbf{Text:} 12 oz bottle poured into a pint glass - a - pours a transparent , pale golden color . the head is pale white with no cream , one finger 's height , and abysmal retention . i looked away for a few seconds and the head was gone \  \underline{s \textbf{\textcolor{red}{-}} stale cereal grains dominate . hardly any other notes to speak of . very mild in strength} t - sharp corn/grainy notes throughout it 's entirety . $\cdots \cdots.$
		 \\
 
		\hline\hline

\end{tabularx}
\vspace{-10pt}
\label{tab:examples}
\end{table*}

In summary, our contributions are:
\begin{itemize}
    \item To the best of our knowledge, we are the first to identify the degeneration problem as a form of spurious correlation. Leveraging probabilistic graphical models, we are the first to comprehensively elucidate feature correlation and degeneration under a unified perspective.  
    \item We find that the target label and non-causal features are \emph{d-separated} by the direct causal features. Based on this insight, we propose the MCD criterion, which opens a new avenue for discovering causal rationales, marking the main contribution of this study. Unlike previous methods, MCD-based methods do not require prior expert knowledge about non-causal features, thus presenting potential for broader applicability.    
    \item We present a simple and practical architecture to develop an MCD-based method. Experiments across various datasets demonstrate that our approach achieves an improvement of up to $13.7\%$ in F1 score compared to state-of-the-art MMI-based rationalization methods.
\end{itemize}

\section{Related work}\label{sec:related}

\textbf{Rationalization}. The basic cooperative framework of rationalization called RNP \citep{rnp} is flexible and offers a unique advantage: certification of exclusion, which means any unselected input is guaranteed to have no contribution to the prediction \citep{interlocking}. Based on this cooperative framework, many methods have been proposed to improve RNP from various aspects. 
\citet{2018rationalegumble} used 
Gumbel-softmax to do the reparameterization for binarized selection. \citet{hardkuma}  replaced the Bernoulli sampling
distributions with rectified Kumaraswamy distributions. \citet{jain2020faith} disconnected the training
regimes of the generator and predictor networks using a saliency threshold. \citet{informationbottle} imposed a discrete bottleneck objective to balance the task performance and the rationale length. \citet{car} tried to select class-wise rationales. \citet{aaai-multiaspect,acl-multiaspect} tried to select rationales belonging to different aspects at once.
\citet{irrationality} called for more rigorous evaluation of rationalization models. \citet{scott}  leveraged meta-learning techniques to improve the quality of the explanations. \citet{cooperative} cooperatively trained the models with standard continuous and discrete optimization schemes. \cite{leakage} explored better metrics for the explanations. \cite{concept} used phrase-based concepts to conduct a self-explaining model. Other methods like data augmentation with pretrained models \citep{counter}, training with human-annotated rationales \citep{Unirex}, have also been tried. These methods are orthogonal to our research.

\textbf{Spurious correlations}.
Several methods have been proposed to address the issues arising from either feature correlation or mask correlation. The impact of feature correlation is somewhat mitigated by techniques such as invariant risk minimization \citep{invarant} or backdoor adjustment \citep{interventional}. However, as indicated in the introduction, these methods have certain limitations. To combat mask correlation, the usual strategy involves introducing an auxiliary module, which has access to the full input, to regulate the original modules and prevent them from overfitting to trivial patterns introduced by the explainer \citep{interlocking,rethinking,liufr}. Other methods like using multiple explainers to select diverse rationales \citep{liumgr}, assigning asymmetric learning rates for the two players \citep{liudr}, have also been tried. Unfortunately, these methods have limited effectiveness against feature correlation in the input data. These aforementioned methods are most relevant to our research, yet we are the first to consider both feature correlation and mask correlation from a unified perspective.

\section{Preliminaries}\label{sec:preliminaries}

We consider the text classification task, where the input is a text sequence  ${X}$=$[x_1,x_2,\cdots,x_l]$ with ${x}_i$ being the $i$-th token and $l$ being the number of tokens. The label of ${X}$ is a one-hot vector $Y\in\{0,1\}^c$, where $c$ is the number of categories. $\mathcal{D}$ represents the training set. Ante-hoc rationalization consists of an explainer $f_E(\cdot)$ and a predictor $f_P(\cdot)$, with $\theta_e$ and $\theta_p$ representing the parameters of the explainer and predictor, respectively. The goal of an MMI-based explainer is to select the most indicative pieces from the input that are related to the label.

For  $(X,Y)\sim\mathcal{D}$, the explainer first outputs a sequence of binary mask $M=f_E(X)=[m_1,\cdots,m_l]\in \{0,1\}^l$ (in practice, the explainer first outputs a Bernoulli distribution for each token and the mask for each token is independently sampled using gumbel-softmax). Then, it forms the rationale candidate $X_Z$ by the element-wise product of $X$ and $M$:
\begin{equation}\label{eqa:getrat}
    X_Z=M\odot X=[m_1x_1,\cdots,m_lx_l].
\end{equation}
To simplify the notation, we denote $f_E(X)$ as $X_Z$ in the following sections, i.e., $f_E(X)=X_Z$. With the generator's selection, we get a set of $(Z,Y)$ pairs, which are generally considered to be samples taken from the distribution $P(Z,Y)$. Then,
vanilla RNP attempts to identify the rationale by maximizing the mutual information $I(Y;X_Z)$:

\begin{equation}
    X_Z^*=\mathop{\arg\max}_{X_Z}I(Y;X_Z)=\mathop{\arg\max}_{X_Z}(H(Y)-H(Y|X_Z))=\mathop{\arg\min}_{X_Z}H(Y|X_Z), \ s.t. \ X_Z=f_E(X).
\end{equation}

In practice, the entropy $H(Y|X_Z)$ is commonly approximated by the minimum cross-entropy $\mathop{\min}_{\theta_p}H_c(Y,\hat{Y}|X_Z)$, with $\hat{Y}=f_P(X_Z)$ representing the output of the predictor. It is essential to note that the minimum cross-entropy is equal to the entropy (please refer to Appendix~\ref{app:relation between entropy and cross-entropy}). Replacing $X_Z$ with $f_E(X)$, the explainer and the predictor are trained cooperatively:
\begin{equation}\label{eqa:overall prediction accuracy}
\mathop{\min}_{\theta_e, \theta_p}H_c(Y,f_P(f_E(X))|f_E(X)), \ s.t., \ (X,Y)\sim \mathcal{D}.
\end{equation}

To make the selected rationale human-intelligible, rationalization methods usually constrain the rationales by compact and coherent regularization terms. In this paper, we use the same constraints used in INVRAT \citep{invarant}:
\begin{equation}\label{eqa:sparse regular}
\Omega (M) = \lambda_1 \bigg \lvert \frac{||M||_1}{l}-s \bigg\rvert +\lambda_2\sum_{t=2}^{l} \big|m_t-m_{t-1} \big|. 
\end{equation} The first term encourages that the percentage of the tokens being selected as rationales is close to a pre-defined level $s$. The second term encourages the rationales to be coherent.

\section{Method}\label{sec:method}

\subsection{Motivation: how spurious correlations come into being.}
In this section, we consider $X$ as a set of variables (or a multi-dimensional variables), and the selected rationale candidate $X_Z$ is a subset (some dimensions) of it.

\begin{wrapfigure}[7]{R}{0.5\columnwidth}
    \vspace{-12pt}
    \includegraphics[width=0.5\columnwidth]{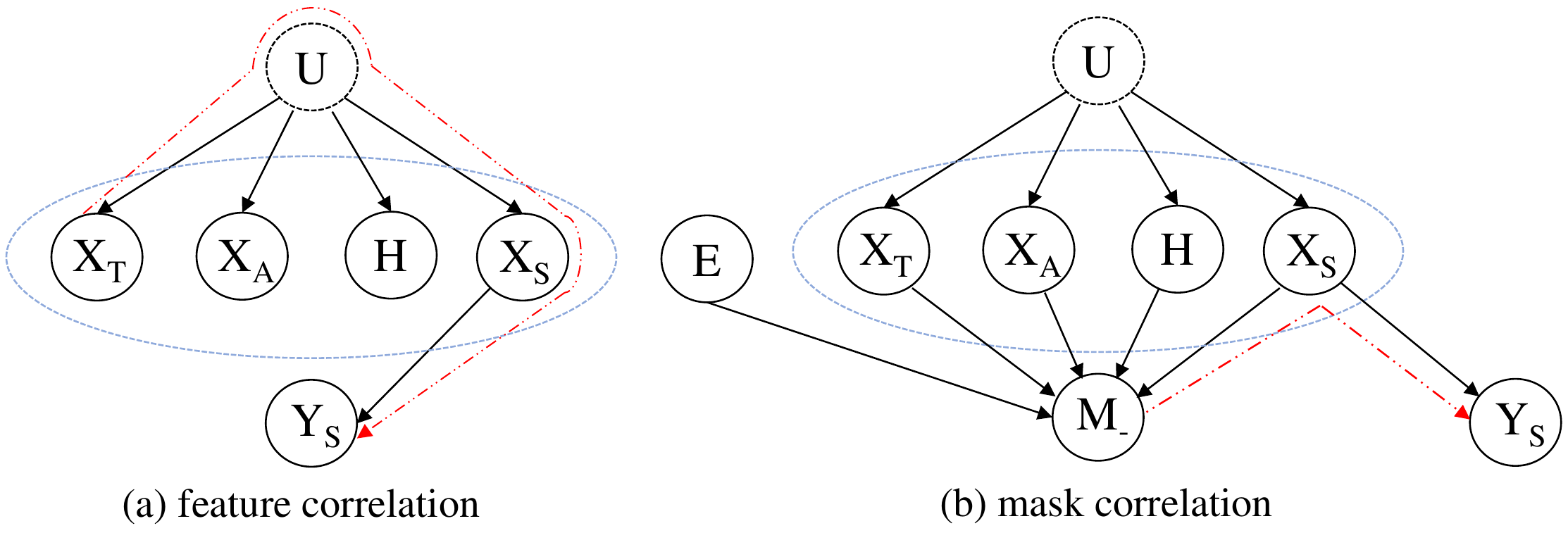}
  \caption{A probabilistic graph for (a) the data generating process of \emph{BeerAdvocate} and (b) the rationale (mask $M$) selection process.}
  \label{fig:causal}
\end{wrapfigure}

To begin with, in Figure~\ref{fig:causal}(a), we posit a probabilistic graphical model to illustrate the corresponding data-generating process for the \emph{BeerAdvocate} dataset. The input $X$ comprises comments on three aspects: $X_S$ for \textbf{S}mell or Aroma, $X_T$ for \textbf{T}aste, and $X_A$ for \textbf{A}ppearance, each of which can be considered as a subset variables of $X$. 
Additionally, $H$ signifies something that does not discuss the sentiment tendency of $X$. For instance, $H$ could include the color of a bottle. 
The annotators assign the smell label $Y_S$ by viewing the comments on aroma ($X_S\xrightarrow{}Y_S$). Therefore, only $X_S$ serves as the direct cause for $Y_S$. However, $X_S$ is correlated with $X_T$ due to a set of unobserved variables $U$ (called  \emph{confounders}). 
For example, $U$ may include a variable indicating whether the beer originates from a reputable brand, and a pleasant taste may imply that the beer comes from a good brand ($U\xrightarrow{} X_T$). Moreover, a beer from a reputable brand is likely to have a pleasing smell ($U\xrightarrow{} X_S$). Consequently, $X_T$ is associated with $Y_S$ via a \emph{backdoor} path, as depicted by the red dotted line in Figure~\ref{fig:causal}(a). In this situation, $X_S$ is somewhat indicative of $Y_S$, but it signifies a statistical correlation rather than causality.

To have a more intuitive understanding of this correlation, we assume a toy example where $U$, $X_S$, $X_T$, and $Y_S$ are all Bernoulli variables, with their respective probability distributions as:
\begin{equation}\label{eqa:assume distribution of figure 2}
    \begin{aligned}
        & p(U=1) = p(U=0) = 0.5, \\
        &p(X_T=1|U=1)=p(X_T=0|U=0)=0.9,\\
        &p(X_S=1|U=1)=p(X_S=0|U=0)=0.9,\\
        &p(Y_S=1|X_S=1)=p(Y_S=0|X_S=0)=0.9,
    \end{aligned}
\end{equation}

With some simple derivations, we can easily obtain (detailed derivation is in Appendix~\ref{proof:other bernoulli}): 
\begin{equation}\label{eqa:other bernoulli}
    p(X_S=1) = p(X_T=1) = p(Y_S=1) = 0.5.
\end{equation}
Then, we can further get (see Appendix~\ref{proof:correlated_aspects} for the detailed derivation of Equation~\ref{eqa:correlated_aspects} and~\ref{eqa:taste2smell}):
\begin{equation}\label{eqa:taste to confounder}
    p(U=1|X_T=1)=\frac{p(U=1,X_T=1)}{p(X_T=1)}=\frac{p(X_T=1|U=1)p(U=1)}{p(X_T=1)}=0.9.
\end{equation}
\begin{equation}\label{eqa:correlated_aspects}
    p(X_S=1|X_T=1)=\sum_{U\in\{0,1\}}p(X_S=1|U)p(U|X_T=1)=0.9*0.9+0.1*0.1=0.82.
\end{equation}
\begin{equation}\label{eqa:taste2smell}
    p(Y_S=1|X_T=1)=\sum_{X_S\in\{0,1\}}p(Y_S=1|X_S)p(X_S|X_T=1)=0.82*0.9+0.18*0.1=0.756.
\end{equation}

Equation~\ref{eqa:correlated_aspects} demonstrates that $X_T$ (Taste) is highly correlated with $X_S$ (Smell), and Equation~\ref{eqa:taste2smell} indicates that $X_T$ (Taste) is also strongly indicative of $Y_S$ (Smell label). This situation can result in numerous local optima during the rationalization training process. Note that becoming trapped in a local optimum poses a significant challenge in rationalization \citep{interlocking,liufr}. 
It is worth noting that the correlation between taste and smell here is merely one of the examples, and sometimes $H$ can also correlate with $Y$ in a similar fashion. For instance, in LIME \citep{LIME}, a predictor is trained to determine whether an image contains a wolf or not based on the presence of snow in the background.

Furthermore, the degeneration problem can also be interpreted with a kind of spurious correlation (called mask correlation), as illustrated in Figure~\ref{fig:causal}(b) with an example. Returning to the second example in Table~\ref{tab:examples}, the variable $M_{\_}$, denoting whether $``-"$ is selected as part of the rationale candidate, is caused by the input $X$ (comprising subsets $X_A$, $X_T$, $X_S$ and $H$) and the explainer $E$. $M_{\_}$ is also correlated with $Y_S$ through a backdoor path, as indicated by the red line in Figure~\ref{fig:causal}(b).

\subsection{The conditional independent property}

We first introduce an important concept in probabilistic graphical models, namely \textbf{d-separation}. Subsequently, we demonstrate how d-separation contributes to the identification of causal rationales.

\textbf{D-Separation} \citep{PRML-2006}: 
$A$, $B$, and $C$ denote arbitrary, non-intersecting sets of nodes (and their union might not cover all nodes of the graph) in a given probabilistic graph. Our objective is to determine whether a specific conditional independence statement $A\upmodels B|C$ is implied by this graph. To do so, we examine all possible paths from any node in $A$ to any node in $B$. A path is said to be blocked if it includes a node $o$ such that either (see Appendix~\ref{app: example of probabilistic graph} for why such a path is blocked)
\begin{itemize}
\item (a) The arrows on the path meet at node $o$, forming either a chain (i.e., $\xrightarrow{} o \xrightarrow{}$) or a fork (i.e., $\xleftarrow{} o \xrightarrow{}$), with the node $o$ being part of set C, or
\item (b) The arrows on the path meet at node $o$ to form a collider (i.e., $\xrightarrow{} o \xleftarrow{}$), and neither the node $o$ itself nor any of its descendants are included in set C.
\end{itemize}
If all paths are blocked, then $A$ is considered to be \textbf{d-separated} from $B$ by $C$, meaning that $A\upmodels B|C$.

Returning to our rationalization problem, the backdoor path (dotted red line) in Figure~\ref{fig:causal}(a) comprises a fork ($X_T\xleftarrow{}U\xrightarrow{}X_S$) and a chain ($U\xrightarrow{}X_S\xrightarrow{}Y_S$). If either $X_S$ or $U$ is included in the conditioning set, the path between $X_S$ and $Y_S$ becomes blocked, leading to their conditional independence, and consequently, the eradication of corresponding feature correlation. Similarly, the backdoor path (dotted red line) in Figure~\ref{fig:causal} (b) forms a fork ($M_{-}\xleftarrow{}X_S\xrightarrow{}Y_S$). By including $X_S$ in the conditioning set, the path between $M_{-}$ and $Y_S$ is blocked, resulting in their conditional independence and consequently, the elimination of corresponding mask correlation.

We consider the general case, where the input $X$ is a set of variables (or features). $X_R$ is a subset of $X$ that exclusively contains all the direct causes of the target label $Y$, i.e., the desiderata of the rationale. We select a subset of $X$ to serve as the rationale candidate (denoted as $X_Z$), while the remaining unselected part is referred to as $X_{-Z}$. This leads us to the following properties:

\begin{lemma}\label{lem:d-sep to causal}
   If $X_{-Z}$ and $Y$ are d-separated by $X_Z$, we then have that all of the direct causal features in $X$ must be included in $X_Z$: 
   \begin{equation}
       X_{-Z}\  \text{and}\  Y \ \text{are d-separated by} \ X_Z \Longrightarrow X_R \subset X_Z.
   \end{equation}
\end{lemma}
The proof is in Appendix~\ref{app:proof of d-sep to causal}. And it's very easy to intuitively understand it: a direct cause has a one-hop path to the label. To block this path, this cause must be included in $X_Z$.  

\begin{assumption}\label{assumption:no causal from y to x}
    The label $Y$ has no causal effect on any variables in $X$.
\end{assumption}

Assumption~\ref{assumption:no causal from y to x} is naturally valid in most real world applications due to the temporal sequence between $X$ and $Y$. We also provide some failure cases of this assumption in Appendix~\ref{app: failure of ass1}. Assumption~\ref{assumption:no causal from y to x} specifies that there is no arrow pointing from $Y$ to any nodes in $X$.

\begin{lemma}\label{lem:causal to d-sep}
    If Assumption~\ref{assumption:no causal from y to x} holds, we have:
    \begin{equation}
         X_{-Z}\  \text{and}\  Y \ \text{are d-separated by} \ X_Z \Longleftarrow X_R \subset X_Z.
    \end{equation}
\end{lemma}

The proof is in Appendix~\ref{app: proof of causal to d-sep}. Combining Lemma~\ref{lem:d-sep to causal} and Lemma~\ref{lem:causal to d-sep}, we then have:

\begin{theorem}\label{the:d-separation for causal}

If Assumption~\ref{assumption:no causal from y to x} holds, then all the direct causal features to $Y$ within $X$ will be included in $X_Z$ if and only if $X_{-Z}$ and $Y$ are d-separated by $X_Z$:

\begin{equation}
         X_{-Z}\  \text{and}\  Y \ \text{are d-separated by} \ X_Z \Longleftrightarrow X_R \subset X_Z.
    \end{equation}
\end{theorem}

\textbf{Remark}. In light of Theorem~\ref{the:d-separation for causal}, we understand that if we aim to achieve $Y\upmodels X_{-Z}|X_Z$, we will consequently incorporate all direct causes of $Y$ into $X_Z$. It should be noted that the compactness of $X_Z$ is facilitated through the sparsity constraint expressed in Equation~\ref{eqa:sparse regular}.

\begin{wrapfigure}[11]{R}{0.55\columnwidth}
\includegraphics[width=0.55\columnwidth]{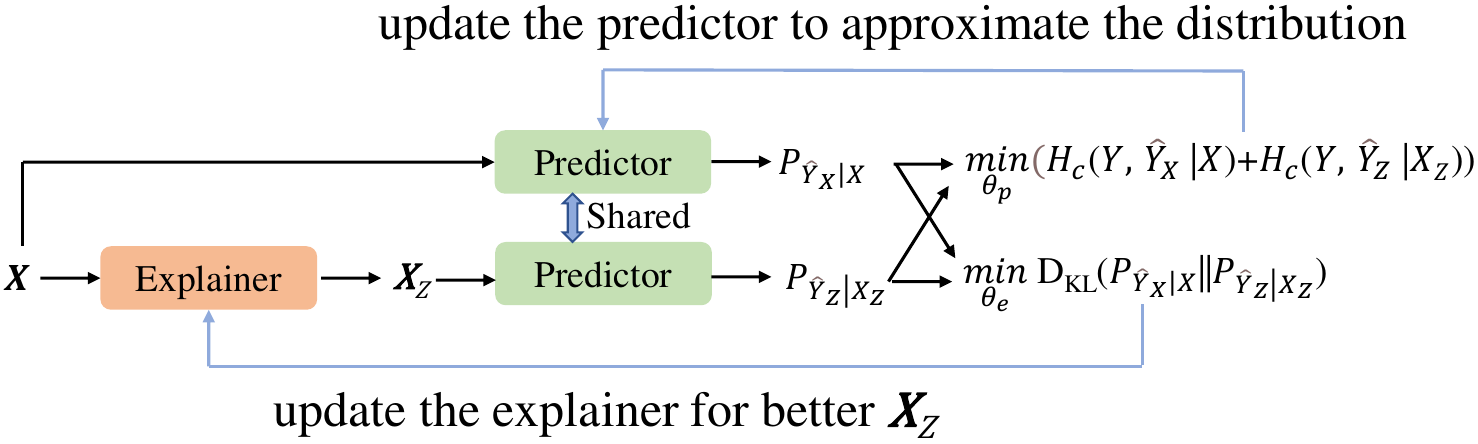}
\caption{The architecture of our proposed MCD. The approximators for the two distributions are shared to reduce the model complexity.}
    \label{fig:proposed bcr}
\end{wrapfigure}

\subsection{The proposed method}
\textbf{Minimum conditional dependence criterion}. Although previous research tried to design various auxiliary modules or regularizers to fix the problems of maximum mutual information criterion \citep{invarant,interventional,liufr}, we do not follow them to move on this line. Based on Theorem~\ref{the:d-separation for causal}, we propose a distinct criterion for identifying the causal rationale, which involves minimizing the dependence between $Y$ and the unselected input $X_{-Z}$, conditioned on $X_Z$:
\begin{equation}\label{eqa:min conditional dependence}
    X_Z^*=\mathop{\arg\min}_{X_Z}\mathcal{C}(Y,X_{-Z}|X_Z),
\end{equation}
where $\mathcal{C}$ is a criterion for dependence. For instance, $\mathcal{C}$ could take the form of partial correlation (applicable only to linear associations), mutual information, divergence, or the Hilbert-Schmidt Independence Criterion (HSIC) \citep{hsic}, among others. 

It then leads to the question of how we can apply this criterion in practice. In this study, we only present a straightforward and practical method to validate our assertion with respect to Theorem~\ref{the:d-separation for causal}, leaving the exploration of other measurements for future work.
We first rewrite $Y\upmodels X_{-Z}|X_Z$ as  
\begin{equation}
    P(Y|X_Z)=P(Y|X_Z,X_{-Z})=P(Y|X).
\end{equation}
Obviously, $P(Y|X_Z)=P(Y|X)$ if and only if the divergence between the two distributions is zero: 
\begin{equation}
    Y\upmodels X_{-Z}|X_Z \iff P(Y|X_Z)=P(Y|X) \iff D_{KL}(P({Y|X})||P({Y|X_Z}))=0.
\end{equation}

\textbf{Estimating divergence through approximation}.
The real distributions of $P(Y|X_Z)$ and $P(Y|X)$ are not directly accessible. So we need further efforts to approximate them. We try to approximate them by making use of the predictor. We first approximate $P(Y|X_Z)$ with $P(\hat{Y}_Z|X_Z)$ by minimizing the cross-entropy $H_c(Y,\hat{Y}_Z|X_Z)$, and we also approximate $P(Y|X)$ with $P(\hat{Y}_X|X)$ by minimizing $H_c(Y,\hat{Y}_X|X)$, where $\hat{Y}_Z,\hat{Y}_X$ are the predictor's outputs with the inputs being $Z$ and $X$, respectively.

Thus, the training process for our MCD is depicted in Figure~\ref{fig:proposed bcr}: the explainer first generates a rationale candidate $X_Z$ from the input $X$. Subsequently, $X_Z$ and $X$ are fed into the predictor to obtain two distributions, $P(\hat{Y}_Z|X_Z)$ and $P(\hat{Y}_X|X)$. By replacing $X_Z$ with $f_E(X)$ and $\hat{Y}$ with $f_P(\cdot)$, the overall objective of our model becomes (The pytorch implementation is in Appendix~\ref{app: pytorch}):
\begin{equation}\label{eqa: overall objective}
\begin{aligned}
    &\mathop{\min}_{\theta_p}E_{(X,Y)\sim \mathcal{D}}[H_c(Y,\hat{Y}_Z|X_Z)+ H_c(Y,\hat{Y}_X|X)]\\&+\mathop{\min}_{\theta_e}E_{(X,Y)\sim \mathcal{D}}[D_{KL}(P(\hat{Y}_X|X)||P(\hat{Y}_Z|X_Z))+\Omega (M)], \\
    & s.t., \ X_Z=f_E(X), \ P(\hat{Y}_X|X)=f_P(X), \ P({\hat{Y}_Z|X_Z})=f_P(X_Z).
\end{aligned}
\end{equation}
Notably, although the first term $H_c(Y,\hat{Y}_Z|X_Z)$ is similar to the one used in Equation~\ref{eqa:overall prediction accuracy}, it is detached from the explainer's parameters $\theta_e$. It is now only used to help the predictor approximate the real distribution $P(Y|X_Z)$ rather than to guide the explainer to find a good rationale.

\section{Experiments}
\subsection{Datasets and metrics}

\textbf{Datasets}\label{sec:dataset}
1) \textbf{BeerAdvocate} \citep{beer} is a multi-aspect sentiment prediction dataset widely adopted in rationalization studies. Given the high correlation among the rating scores of different aspects within the same review, rationale selection encounters severe feature correlation challenges.
Following INVRAT \citep{invarant} and Inter\_RAT \citep{interventional}, we utilize the original dataset (which we refer to as \emph{correlated BeerAdvocate}) to verify MCD's effectiveness in handling both feature correlation and mask correlation simultaneously. 2) \textbf{HotelReviews} \citep{hotel} is another multi-aspect sentiment classification dataset containing less feature correlation, which is used by the latest SOTA method FR \citep{liufr} to evaluate the effectiveness of addressing degeneration. We utilize the \emph{Service} aspect to further demonstrate the competitive edge of our MCD. Among these datasets, each aspect itself can be seen as a dataset and is trained independently.

\textbf{Metrics}. Considering that the annotators assign the label of the target aspect by observing the causal features, the overlap between the tokens selected by the model and those annotated by humans provides a robust metric for rationale causality. The terms $P, R, F1$ denote precision, recall, and $F1$ score respectively. These metrics are the most frequently used in rationalization. The term $S$ represents the average sparsity of the selected rationales, that is, the percentage of selected tokens in relation to the full text. $Acc$ stands for the predictive accuracy.

\subsection{Baselines and implementation details}

We compare with various recent MMI-based methods that are highly relevant to our study. These include methods like INVRAT~\citep{invarant} and Inter\_RAT~\citep{interventional}, which are focused on addressing feature correlation, as well as methods such as FR~\citep{liufr} that aim to mitigate mask correlation (i.e., degeneration). Among these, FR represents the latest SOTA approach in addressing mask correlation, while Inter\_RAT stands as the SOTA in handling feature correlation.

Both the explainer and the predictor are composed of an encoder (which can be an RNN or Transformer) and a linear layer.
Some of the baseline methods have not provided runnable source codes. To ensure a fair comparison, we keep the major settings consistent with those of the baselines, which are commonly utilized in the field of rationalization~\citep{invarant,interlocking,liufr,interventional}. Specifically, we employ the 100-dimensional GloVe~\citep{glove} for word embedding and 200-dimensional GRUs~\citep{gru} to obtain text representation. 
The re-parameterization trick for binarized selection is Gumbel-softmax~\citep{2016gumble}. The hyperparameters of the reimplemented baselines are initialized with the values reported in their source codes, and are then manually tuned multiple times to determine the optimal settings. 
We do not use BERT~\citep{bert} in the main experiments because some recent research \citep{danqi,liufr,cr} has found it to be a challenging task to fine-tune large pretrained models within the rationalization framework (see Appendix~\ref{app:bert} for more discussion). However, as a supplement, we also conduct experiments with two pretrained models, ELECTRA~\citep{electra} and BERT.
The optimizer is Adam~\citep{adam}. All models are trained on a RTX3090 GPU. More details are in Appendix~\ref{app:implemenration details}.

\begin{table*}[t]
    \centering
     \caption{Results on \emph{correlated BeerAdvocate}. Each aspect is trained independently. $``*"$: results obtained from Inter\_RAT \citep{interventional}. The second best F1 scores are \underline{underlined}.}
    \resizebox{0.99\columnwidth}{!}{
    \begin{tabular}{c c c |r r| r r r|r r| r r r|r r| r r r }
\hline
\multicolumn{3}{c|}{\multirow{2}{*}{Methods}} & \multicolumn{5}{c|}{Appearance} & \multicolumn{5}{c|}{Aroma} & \multicolumn{5}{c}{Palate}\\
\cline{4-18}
\multicolumn{3}{c|}{} &\multicolumn{1}{c}{S}& \multicolumn{1}{c|}{Acc} & \multicolumn{1}{c}{P} & \multicolumn{1}{c}{R} &\multicolumn{1}{c|}{F1} &\multicolumn{1}{c}{S}& \multicolumn{1}{c|}{Acc} & \multicolumn{1}{c}{P} & \multicolumn{1}{c}{R} &\multicolumn{1}{c|}{F1} &\multicolumn{1}{c}{S}& \multicolumn{1}{c|}{Acc}& \multicolumn{1}{c}{P} & \multicolumn{1}{c}{R} &\multicolumn{1}{c}{F1}\\
\hline
\multicolumn{3}{c|}{RNP$^{*}$} &10.0&\multicolumn{1}{c|}{-}& 32.4 & 18.6 & 23.6  &10.0&\multicolumn{1}{c|}{-}&44.8 &32.4& 37.6 & 10.0 & \multicolumn{1}{c|}{-} & 24.6 & 23.5 & 24.0\\
\multicolumn{3}{c|}{INVRAT$^{*}$} & 10.0 & \multicolumn{1}{c|}{-} & 42.6 &31.5 & 36.2 &10.0&\multicolumn{1}{c|}{-} & 41.2 & 39.1 & 40.1&10.0&\multicolumn{1}{c|}{-}&34.9&45.6&39.5\\
\multicolumn{3}{c|}{Inter-RAT$^{*}$} & 11.7 & \multicolumn{1}{c|}{-}&66.0 &46.5 & \underline{54.6} &11.7 & \multicolumn{1}{c|}{-} & 55.4 & 47.5&51.1&12.6&\multicolumn{1}{c|}{-}&34.6&48.2&40.2\\
\multicolumn{3}{c|}{FR} & 11.1 & 75.8&70.4 &42.0 & {52.6} &9.7 & 87.7 & 68.1 & 42.2&\underline{52.1}&11.7&87.9&43.7&40.9&\underline{42.3}\\
\multicolumn{3}{c|}{MCD(ours)} &9.5&81.5&\textbf{94.2}&\textbf{48.4}&\textbf{63.9}&9.9&87.5&\textbf{84.6}&\textbf{53.9}&\textbf{65.8}&9.4&87.3&\textbf{60.9}&\textbf{47.1}&\textbf{53.1}
  \\
  \hline
  \hline
  \multicolumn{3}{c|}{RNP$^{*}$} &20.0&\multicolumn{1}{c|}{-}& 39.4 & 44.9 & 42.0  &20.0&\multicolumn{1}{c|}{-}&37.5&51.9&43.5  & 20.0 & \multicolumn{1}{c|}{-}& 21.6 & 38.9 & 27.8\\
\multicolumn{3}{c|}{INVRAT$^{*}$} & 20.0 & \multicolumn{1}{c|}{-} & 58.9 &67.2 & 62.8 &20.0&\multicolumn{1}{c|}{-} & 29.3 & 52.1 & 37.5&20.0&\multicolumn{1}{c|}{-}&24.0&55.2&33.5\\
\multicolumn{3}{c|}{Inter-RAT$^{*}$} & 21.7 & \multicolumn{1}{c|}{-}&62.0 &76.7 & 68.6 &20.4 & \multicolumn{1}{c|}{-} & 44.2 & 65.4&52.8&20.8&\multicolumn{1}{c|}{-}&26.3&59.1&36.4\\
\multicolumn{3}{c|}{FR} & 20.9 & 84.6&74.9&84.9 & \underline{79.6} &19.5 & 89.3 & 58.7 &73.3&\underline{65.2}&20.2&88.2&36.6&59.4&45.3\\
\multicolumn{3}{c|}{MCD(ours)} &20.0&85.5&\textbf{79.3}&\textbf{85.5}&\textbf{82.3}&19.3&88.4&\textbf{65.8}&\textbf{81.4}&\textbf{72.8}&19.6&{87.7}&\textbf{41.3}&\textbf{65.0}&\textbf{50.5}
  \\
  \hline
  \hline
  \multicolumn{3}{c|}{RNP$^{*}$} &30.0&\multicolumn{1}{c|}{-}& 24.2 & 41.2 & 30.5  &30.0&\multicolumn{1}{c|}{-}&27.1 &55.7& 36.4& 30.0 &\multicolumn{1}{c|}{-}& 15.4 & 42.2 & 22.6 \\
\multicolumn{3}{c|}{INVRAT$^{*}$} & 30.0 & \multicolumn{1}{c|}{-} & 41.5 &74.8 & 53.4 &30.0&\multicolumn{1}{c|}{-} & 22.8 & 65.1 & 33.8&30.0&\multicolumn{1}{c|}{-}&20.9&{71.6}&32.3\\
\multicolumn{3}{c|}{Inter-RAT$^{*}$} & 30.5 & \multicolumn{1}{c|}{-}&48.1 &82.7 & 60.8 &29.4 & \multicolumn{1}{c|}{-} & 37.9 & 72.0&49.6&30.4&\multicolumn{1}{c|}{-}&21.8&{66.1}&{32.8}\\
\multicolumn{3}{c|}{FR} & 29.6 & 86.4&50.6 &81.4 & \underline{62.3}&30.8 & 88.1 & 37.4 & 75.0&49.9&30.1&87.0&24.5&58.8&\underline{34.6}\\
\multicolumn{3}{c|}{MCD(ours)} &29.7&86.7&\textbf{59.6}&\textbf{95.6}&\textbf{73.4}&29.6&90.2&\textbf{46.1}&\textbf{87.5}&\textbf{60.4}&29.4&{87.0}&\textbf{30.5}&\textbf{72.4}&\textbf{42.9}
  \\
  \hline
\end{tabular}
}
\vspace{-12pt}  
    \label{tab:correlated beer}
\end{table*}

\subsection{Results}

\begin{wraptable}[8]{t}{0.45\columnwidth}
\vspace{-18pt}
 \caption{Results on \emph{HotelReview}. $``$*$"$: results obtained from FR \citep{liufr}.}
    \centering
    \resizebox{0.45\columnwidth}{!}{
    \begin{tabular}{c c c |r r| r r r}
    \hline
\multicolumn{3}{c|}{{Methods}}  &\multicolumn{1}{c}{S}& \multicolumn{1}{c|}{Acc} & \multicolumn{1}{c}{P} & \multicolumn{1}{c}{R} &\multicolumn{1}{c}{F1}\\
\hline
         \multicolumn{3}{c|}{RNP*}&11.0 &97.5 & 34.2 & 32.9&33.5 \\
         \hline
         \multicolumn{3}{c|}{DMR*} & 11.6 & \multicolumn{1}{c|}{-} & 43.0 &43.6&43.3\\
         \hline
         \multicolumn{3}{c|}{A2R*} &11.4 & 96.5 & 37.3 & 37.2&37.2\\
         \hline
         \multicolumn{3}{c|}{FR*}&11.5& 94.5&{44.8}&{44.7}&\underline{44.8}\\
         \hline
         \multicolumn{3}{c|}{MCD(ours)}&11.8&97.0 &\textbf{47.0}&\textbf{48.6}&\textbf{47.8}\\
         
         \hline
    \end{tabular}
   }
    \label{tab:hotel results}
\end{wraptable}

\begin{figure}[t]
    \centering
    \includegraphics[width=0.99\columnwidth]{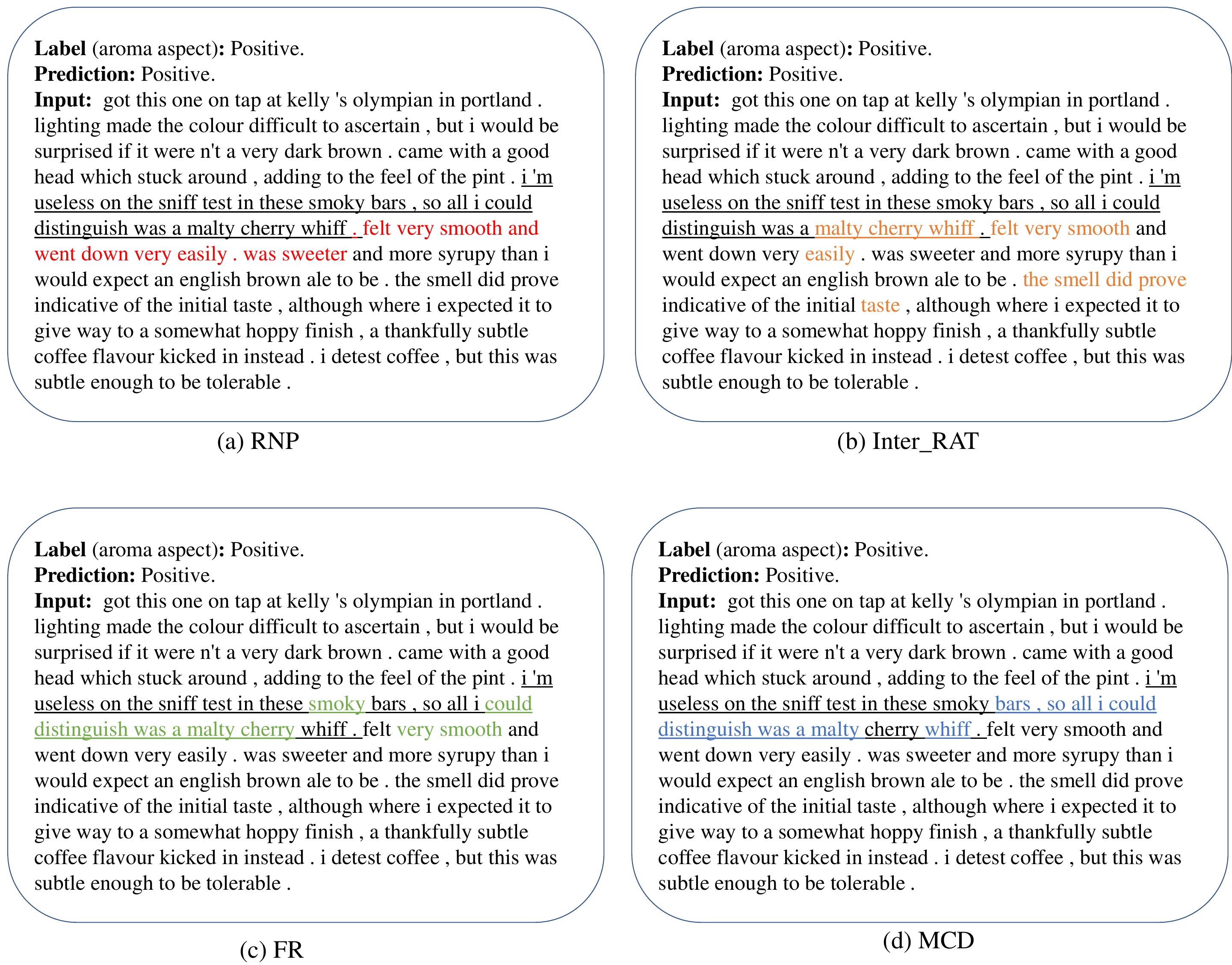}
    \caption{An example of selected rationales in the \emph{Aroma} aspect of BeerAdvocate. The sparsity is set to be about $10\%$. The \underline{underlined} texts are human-annotated rationales. (a): RNP selects palate only. (b): Inter\_RAT selects aroma but also palate ($``$felt very smooth$"$). (c): FR is similar to Inter\_RAT. (d): MCD selects aroma only.}
    \label{fig:good example}
\end{figure}

\textbf{Comparison with SOTA Methods}. Table~\ref{tab:correlated beer} shows the results on \emph{correlated BeerAdvocate} with the rationale sparsity being about $10\%$, $20\%$, and $30\%$.  We set the sparsity to be similar to previous methods by adjusting the sparsity regularization term (i.e., $s$) in Equation~\ref{eqa:sparse regular}. Compared to MMI-based methods, we gain significant improvements across all three aspects and three different sparsity. In particular, we improve the F1 score by more then $10\%$ as compared to the previous SOTA in three settings: in the \emph{Aroma} aspect with $S\approx 10$, the \emph{Palate} aspect with $S\approx 10$, and the \emph{Appearance} aspect with $S\approx 30$. We show an visualized example of the selected rationales in Figure~\ref{fig:good example}.
Since our MCD criterion (Equation~\ref{eqa:min conditional dependence}) is not limited to a specific measurement of dependence, we also conduct experiments by replacing KL-divergence with JS-divergence, and the results are in Appendix~\ref{app: experiments with js}.
Table~\ref{tab:hotel results} shows the results on another dataset also used in FR, where DMR \citep{dmr} and A2R \citep{interlocking} are two recent MMI-based methods. For this dataset, we follow FR to set the sparsity similar to that of the human-annonated rationales. On this dataset, we still beat all the MMI-based methods. 
We also show the time efficiency in Appendix~\ref{app:time efficiency}.

\textbf{Inducing mask correlation with skewed explainer}. 
In order to evaluate scenarios where feature correlation is not severe and our primary concern is mask correlation, we follow FR's approach to conduct experiments in a synthetic setting where the explainer is specifically initialized to induce mask correlation, also referred to as degeneration.
The details of the initialization can be found in Appendix~\ref{app:skew explainer}. Following FR, we utilize the \emph{Palate} aspect of \emph{decorrelated BeerAdvocate} dataset (a subset of the original \emph{BeerAdvocate} that has been filtered by \citet{rnp}). This subset contains less feature correlation compared to the original dataset.
The results are presented in Table~\ref{tab:skew generator}, where skew$k$ and $Pre\_acc$ indicates the degree of mask correlation.
In this situation, the vanilla RNP fails to identify the causal rationales, and FR is also significantly impacted when the degree of mask correlation is high. Our MCD is much less affected, demonstrating its robustness in such scenarios.

\begin{table*}[t]
\caption{Results of skewed explainer that induces degeneration (i.e., mask correlation) in the \emph{Palate} aspect of \emph{BeerAdvocate}. $``*"$: results obtained from the paper of FR.}
    \centering
    \resizebox{0.99\columnwidth}{!}{
    \begin{tabular}{c |c c c| c c c| c c c| c c c| c c c| c c c}
    \hline
    \multicolumn{1}{c|}{\multirow{2}{*}{Setting}} & \multicolumn{6}{c|}{RNP*} & \multicolumn{6}{c|}{FR*}& \multicolumn{6}{c}{MCD(ours)}\\
\cline{2-19}
\multicolumn{1}{c|}{} &Pre\_acc&S&Acc & R & R &F1&Pre\_acc&S&Acc & P & R &F1&Pre\_acc&S&Acc & P & R &F1\\
\hline
\multicolumn{1}{c|}{skew65.0} &66.6&14.0&83.9 &40.3 &45.4&42.7&66.3&14.2&81.5&{59.5}&\textbf{67.9}&\textbf{63.4}&66.3&12.9&84.6&\textbf{61.6}&{63.7}&{62.6}\\
\multicolumn{1}{c|}{skew70.0} &71.3&14.7&84.1&10.0 & 11.7 & 10.8&70.8&14.1&88.3&{54.7}& {62.1} &{58.1}&70.2&13.5&81.1&\textbf{59.0}& \textbf{64.0} &\textbf{61.4}\\
\multicolumn{1}{c|}{skew75.0} &75.5 & 14.7 &87.6 &8.1&9.6&8.8&75.6&13.1&84.8&{49.7}& {52.2} &{51.0}&75.3&13.4&84.2&\textbf{61.3}& \textbf{65.1} &\textbf{63.1}\\
\hline
    \end{tabular}
    }
    
    \label{tab:skew generator}
\end{table*}

\begin{table*}[t]
\caption{Results of methods using pretrained ELECTRA as the encoder. 
}

        \resizebox{0.99\columnwidth}{!}{
\begin{tabular}{c c c |c c| c c c|c c| c c c |c c| c c c }
\hline
\multicolumn{3}{c|}{\multirow{2}{*}{Methods}} & \multicolumn{5}{c|}{Appearance} & \multicolumn{5}{c|}{Aroma} & \multicolumn{5}{c}{Plate}\\
\cline{4-18}
\multicolumn{3}{c|}{} &S& Acc & P & R &\multicolumn{1}{c|}{F1} &S& Acc & P & R &\multicolumn{1}{c|}{F1} &S& Acc& P & R &\multicolumn{1}{c}{F1}\\
\hline
\multicolumn{3}{c|}{FR-ELECTRA} & 16.3&86.5& 19.1 & 17.0 &18.0&14.8 &85.9& 58.6 & 54.8&56.7&11.2&78.0&12.0&10.7&11.3\\
\multicolumn{3}{c|}{MCD-ELECTRA} &18.5&90.0&\textbf{84.8} & \textbf{85.6}&\textbf{85.2}&14.5&86.6& \textbf{86.2}&\textbf{78.7}&\textbf{82.3}&12.1&85.0&\textbf{63.0}&\textbf{60.3}&\textbf{61.6}
  \\\hline
\end{tabular}
}

\label{tab:DR AND FR with electra}
\vspace{-5pt}
\end{table*}

\begin{wraptable}[10]{T}{0.4\columnwidth}
 \vspace{-19pt}
 \caption{The  F1 scores of models trained with different encoders. $``$*$"$: results obtained from \citep{danqi}. $``$**$"$: results obtained from FR.
 The dataset is decorrelated \emph{Beer-Appearance}.}

    \centering
    \resizebox{0.35\columnwidth}{!}{
    \begin{tabular}{c|c|c|c}
    \hline
         Method& GRU & ELECTRA&BERT  \\
         \hline
         VIB*& -&-&20.5 \\
         \hline
         SPECTRA* & -&-&28.6\\
         \hline
         RNP**&72.3&13.7&14.7\\
         \hline
         FR**&82.8&14.6&29.8\\
         \hline
         MCD(ours) &80.1&85.2&87.1\\
         \hline
    \end{tabular}
   }
    \label{tab:models with bert}
\end{wraptable}

\textbf{Experiments with pretrained language models}. In the field of rationalization, researchers generally focus on frameworks of the models and the methodology rather than engineering SOTA. The methods most related to our work do not use BERT or other pre-trained encoders \citep{invarant, interlocking,liufr,interventional}. Experiments in some recent work 
\citep{danqi,liufr}
suggest that there are some unforeseen obstacles making it hard to finetune large pretrained models within the rationalization framework. For example, Table~\ref{tab:models with bert} shows that two improved rationalization methods (VIB \citep{informationbottle} and SPECTRA \citep{spec}) and the latest published FR all fail to find the informative rationales when replacing GRUs with pretrained BERT. To eliminate potential factors that could lead to an unfair comparison, we adopt the most widely used GRUs as the encoders in our main experiments, which can help us focus more on substantiating our claims themselves, rather than unknown tricks. 
But to show the competitiveness of our MCD, we also provide some experiments with pretrained language models as the supplement. Due to limited GPU resources, we adopt the relatively small ELECTRA-small in all three aspects of \emph{BeerAdvocate} and the relatively large BERT-base in the \emph{Appearance} aspect. We compare our MCD with the latest SOTA FR \citep{liufr}. We follow FR to set the sparsity similar to human-annotated rationales. More details are in Appendix~\ref{app:bert}.

The results with BERT are shown in Table~\ref{tab:models with bert} and results with ELECTRA are shown in Table~\ref{tab:DR AND FR with electra}.  We see that our method can greatly benefit from pretrained models.
In fact, recent research has found that finetuning large pretrained models can be easily affected by overfitting \citep{bert-overfit}, and spurious correlations can exacerbate this overfitting, particularly in larger models \citep{irm-overfit,impossible-env}, which somewhat explains the great progress achieved by our MCD.

\section{Conclusion, future work, and limitations}

In this study, we first illustrate the two primary issues of feature correlation and degeneration in MMI-based rationalization under a unified causal perspective. Subsequently, we uncover the conditional independence relationship between the target label and non-causal and causal features. Based on this observation, we propose a criterion of minimizing conditional dependence to concurrently address the two aforementioned problems. 

Given the versatility of the self-explaining rationalization framework, our proposed methods show significant potential for application across diverse fields such as computer vision and graph learning. Additionally, with the recent remarkable success of large language models (LLMs), exploring how our MCD can aid in training  trustworthy LLMs is another avenue worth pursuing.

A potential limitation is that, similar to IRM-based methods, our primary focus is on identifying rationales with causal effects, rather than quantitatively computing the precise values of these causal effects. Although quantifying causality often relies on strong assumptions, this quantification may be a desideratum for certain applications. We aim to explore this direction in future work to accommodate a wider range of applications. Another limitation is that we focus on the text classification task. Different tasks may have very different causal structures. Thus, how to extend this method to other tasks is also a challenge that needs to be explored. The third limitation is that the obstacles in utilizing powerful pretrained language models under the rationalization framework remain mysterious. Although we have made some progress in this direction, we have to say that the
empirical results with pretrained models are very sensitive to hyperparameter tuning. A recent paper has also shown that very small changes in hyperparameters can lead to significant differences in results (see Remark 6.1 and Appendix G.2 in \citep{ccr}). To avoid being distracted by irrelevant factors, until this issue is resolved, we call for research papers to use small models to better verify their claims. 

\section{Acknowledgements}
This work is supported by National Natural Science Foundation of China under grants 62376103, 62302184, 62206102, and Science and Technology Support Program of Hubei Province under grant 2022BAA046. We are also grateful for the valuable suggestions provided by the anonymous reviewers, which greatly helped to improve the quality of this paper.

\clearpage
\bibliographystyle{acl_natbib}
\bibliography{custom}







\clearpage
\appendix

\section{More Results}
\subsection{More implementation details}\label{app:implemenration details}
To the best of our knowledge, both datasets are sufficiently anonymized to make identification of individuals impossible without significant effort. Both datasets are in English. For \emph{correlated BeerAdvocate}, we preprocess the data in the same way as Inter\_RAT \citep{interventional}. For \emph{Hotel Reviews}, we preprocess them in the same way as FR \citep{liufr}. The maximum text length is set to 256. More statistics of the datasets are in Table~\ref{tab:dataset}. The dataset of \emph{BeerAdvocate} is unbalanced. For the training data, we sample from the positive data to get same number of positive and negative texts. 

\begin{table}[t]
   \centering
   \caption{Statistics of datasets used in this paper. *: the decorrelated BeerAdvocate.}
  \resizebox{0.99\columnwidth}{!}{  
    \begin{tabular}{c l| c c| c c| c c c}
    \hline
         \multicolumn{2}{c|}{\multirow{2}{*}{Datasets}}&\multicolumn{2}{c|}{Train}&\multicolumn{2}{c|}{Dev}&\multicolumn{3}{c}{Annotation}  \\
         \multicolumn{2}{c|}{}& Pos&Neg&Pos&Neg&Pos&Neg&Sparsity\\
         \hline\hline
        \multirow{3}{*}{Beer}&Appearance&202385&12897 &28488&1318&923&13&18.5\\
        {}&Aroma&172299&30564&24494&3396&848&29&15.6\\
        {}&Palate&176038&27639&24837&3203&785&20&12.4\\
        \hline\hline
        \multirow{3}{*}{Beer*}&Appearance&16891&16891 &6628&2103&923&13&18.5\\
        {}&Aroma&15169&15169&6579&2218&848&29&15.6\\
        {}&Palate&13652&13652&6740&2000&785&20&12.4\\
        \hline\hline
        \multirow{3}{*}{Hotel}&Location&7236&7236 &906&906&104&96&8.5\\
        {}&Service&50742&50742&6344&6344&101&99&11.5\\
        {}&Cleanliness&75049&75049&9382&9382&99&101&8.9\\
        \hline
    \end{tabular}
    }
    
    \label{tab:dataset}
\end{table}

Some previous methods needs very careful hyper-parameter tuning. To make fair comparisons, most results of the baselines are copied from previous papers. 
 
The early stopping technique is conducted according to the predictive accuracy of the development set.

For \emph{BeerAdvocate}, we use a learning rate of 0.0001 and a batchsize of 128 for our MCD. For \emph{HotelReview}, we use a learning rate of 0.0001 and a batchsize of 256. 

We report the average results of our MDC by running it with five different random seeds.

\subsection{Pytorch implementation of Equation~\ref{eqa: overall objective}}\label{app: pytorch}

For a batch of $(X,Y)$, we first send $X$ to the explainer to get $X_Z$:
\begin{equation}
    X_Z=f_e(X).
\end{equation}
Then we get a copy of $X_Z$ with the pytorch function $``$torch.detach()$"$:
\begin{equation}
    X_Z'=\text{torch.detach}(X_Z).
\end{equation}

Then, we get $\hat{Y}_X$ and $\hat{Y}_Z'$: 
\begin{equation}
\begin{aligned}
    &\hat{Y}_X=f_p(X),\\
    &\hat{Y}_Z'=f_p(X_Z').
\end{aligned}
\end{equation}
Then we update the predictor with
\begin{equation}
    \mathop{\min}_{\theta_p}[\text{torch.nn.functional.cross\_entropy}(\hat{Y}_Z',Y)+ \text{torch.nn.functional.cross\_entropy}(\hat{Y}_X,Y)],
\end{equation}
which is the first part of Equation~\ref{eqa: overall objective}. At the same time, we update the explainer with Equation~\ref{eqa:sparse regular}.

Now, we deal with the second part of Equation~\ref{eqa: overall objective}.
We first freeze the predictor's parameters and get $X_Z$ again:
\begin{equation}
    X_Z=f_e(X).
\end{equation}
We now do not copy $X_Z$. Instead, we directly get $\hat{Y}_X$ and $\hat{Y}_Z$: 
\begin{equation}
\begin{aligned}
    &\hat{Y}_X=f_p(X),\\
    &\hat{Y}_Z=f_p(X_Z).
\end{aligned}
\end{equation}

Then we update the explainer with 
\begin{equation}\label{eqa: torch kl}
    \min_{\theta_e}\text{F.kl\_div}(\text{F.softmax}(\hat{Y}_Z).log(), \text{F.softmax}(\hat{Y}_X)),
\end{equation}
where ``F'' denotes ``nn.functional''. In practice, we have added Equation~\ref{eqa:sparse regular} to \ref{eqa: torch kl}. 

Now, an update round for Equation~\ref{eqa: overall objective} is completed, and we repeat the above steps again.

\subsection{Details of the skewed explainer}\label{app:skew explainer}
We pretrain the explainer separately using the text classification label as the mask label of the first token. In other words, for texts of class 1, we force the explainer to select the first token, and for texts of class 0, we force the explainer not to select the first token. So, the explainer learns the category implicitly by whether the first token is chosen and the predictor only needs to learn this position information to make a correct prediction. 

$k$ in “skew$k$” denotes the threshold of the skew: we pretrain the explainer as a special classifier of the first token for a few epochs until its prediction accuracy is higher than $k$. Since the accuracy increases rapidly in the first a few epochs, obtaining a model that precisely achieves the pre-defined accuracy is almost impossible. So, we use “$Pre\_acc$” to denote the actual prediction accuracy of the explainer-classifier when the pre-training process stops. Higher “$Pre\_acc$” means easier to degenerate.

\subsection{Discussion on BERT encoder}\label{app:bert}

\begin{wraptable}[8]{R}{0.5\columnwidth}
\vspace{-15pt}
   \centering
   \caption{Results with BERT. VIB: \citet{informationbottle}, SPECTRA: \citet{spec}. The results are from Table 4 of \citep{danqi}. The metric is F1 score.}
   \setlength\tabcolsep{2pt}
    \begin{tabular}{c |c| c}
    \hline
    Methods & Beer-Appearance &Hotel-Cleanliness\\
    \hline
    VIB&20.5&23.5\\
    SPECTRA&28.6&19.5\\
    \hline
         
    \end{tabular}
    
    \label{tab:res_bert}
\end{wraptable}

 In the field of rationalization, researchers generally focus on frameworks of the models and the methodology. Methods most related to our work do not use Bert or other pre-trained encoders \citep{invarant,dmr,rethinking,interlocking,interventional}. We use GRUs and GloVe to ensure the same experimental setup as our baselines for a fair comparison.

 More importantly, how to finetune large models on the rationalization framework is still a significant challenge. Some recent studies \citep{danqi} show that the methods with BERT encoders perform much worse than those with simple GRUs on BeerAdvocate and HotelReviews, which is shown in Table~\ref{tab:res_bert}. VIB and SPECTRA are two RNP-based models. When using BERT, these two methods perform much worse than the vanilla RNP with GRUs. Table~\ref{tab:cr bert} shows the results of a recent workshop paper CR \citep{cr}, which are also much worse than those with GRUs.  
 
 \begin{wraptable}[6]{R}{0.5\columnwidth}
 \vspace{-15pt}
   \centering
   \caption{The F1 scores of CR \citep{cr} with pretrained BERT on \emph{BeerAdvocate}. The results are from Table 1 of \citep{cr}.}
   \setlength\tabcolsep{2pt}
    \begin{tabular}{c| c |c| c}
    \hline
    Method & Appearance &Aroma&Palate\\
    \hline
    CR&27.4&39.0&22.6\\
    \hline
         
    \end{tabular}
    
    \label{tab:cr bert}
\end{wraptable} 

We also conduct experiments with pretrained language models and compare with previous methods.
 As previous methods are not designed to address feature correlations in the original dataset, they typically utilize the \emph{decorrelated BeerAdvocate}) dataset where feature correlation is manually filtered by~\citet{rnp}, focusing mainly on mask correlation. Following previous methods \citep{danqi,liufr,cr}, we use the~\emph{decorrelated BeerAdvocate} dataset. And we set the rationale sparsity to be similar to that of human-annotated rationales. The results are in Table~\ref{tab:models with bert} and Table~\ref{tab:DR AND FR with electra}.
 
\begin{wrapfigure}[9]{R}{0.5\columnwidth}
\vspace{-25pt}
\subfigure[Beer-Appearance]{
        \includegraphics[height=2.4cm]{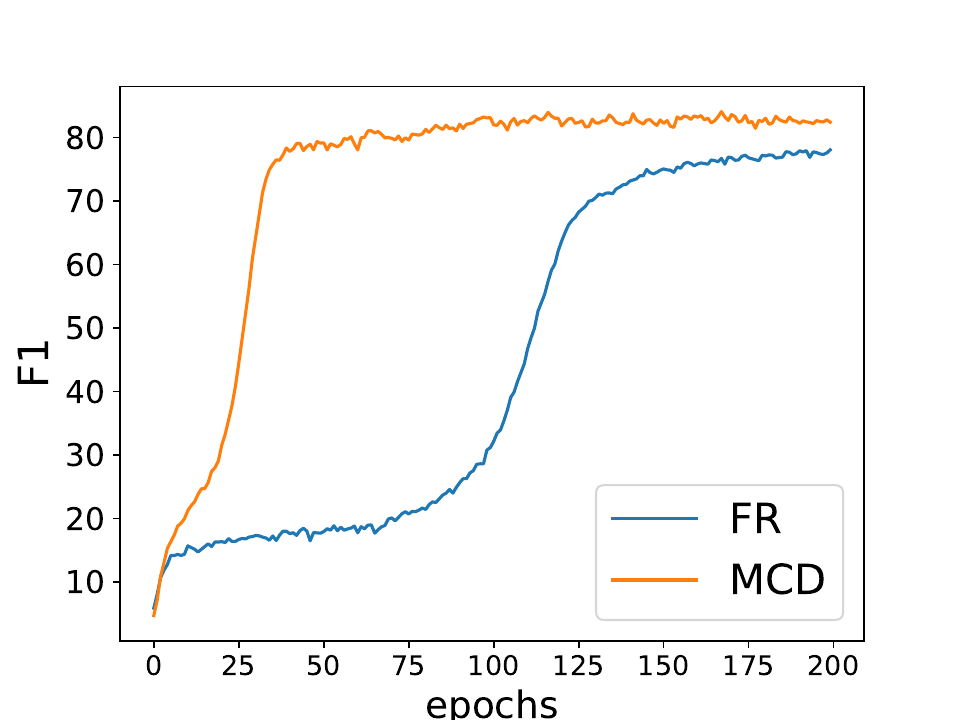}  
    }
\subfigure[Beer-Aroma]{
        \includegraphics[height=2.4cm]{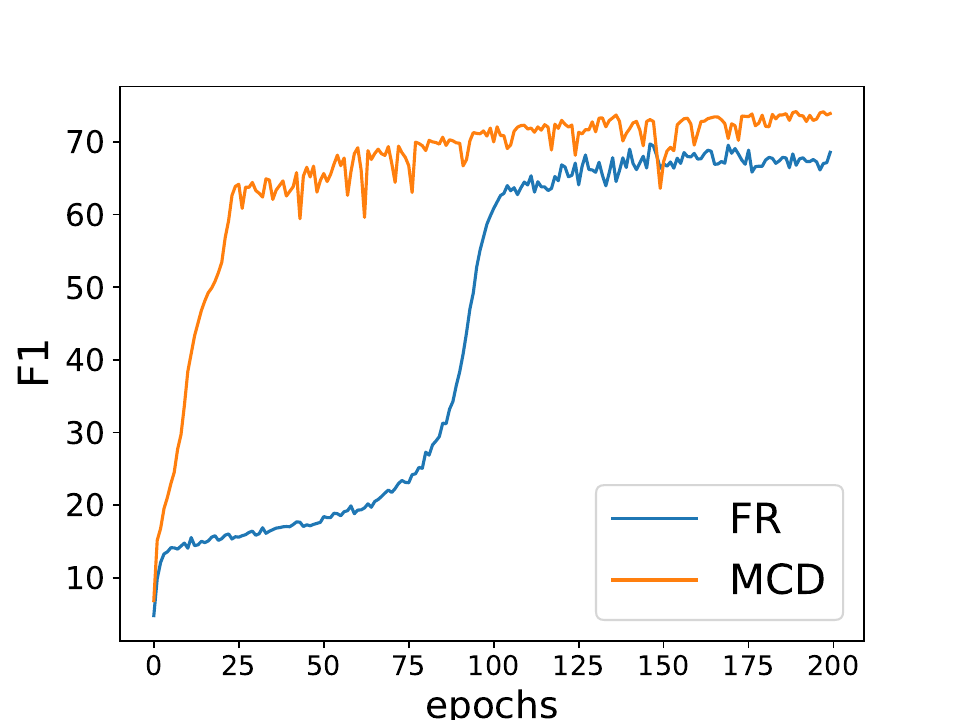}  
    }
  \caption{A comparison of convergence speed between our MCD and the latest MMI-based SOTA FR.}
  \label{fig:convergence speed}
\end{wrapfigure}

\subsection{Experiments with JS-divergence}\label{app: experiments with js}
Since our MCD criterion (Equation~\ref{eqa:min conditional dependence}) is not limited to a specific measurement of dependence, we also conduct experiments by replacing KL-divergence with JS-divergence. The results are in Table~\ref{tab:js}. With either KL-divergence or JS-divergence, our MCD criterion always beat all the MMI-based baselines, showing the effectiveness of MCD.

\begin{table*}[t]
    \centering
     \caption{Results on \emph{correlated BeerAdvocate}. Each aspect is trained independently. $``*"$: results obtained from Inter\_RAT \citep{interventional}. The second best F1 scores are \underline{underlined}.}
    \resizebox{0.99\columnwidth}{!}{
    \begin{tabular}{c c c |r r| r r r|r r| r r r|r r| r r r }
\hline
\multicolumn{3}{c|}{\multirow{2}{*}{Methods}} & \multicolumn{5}{c|}{Appearance} & \multicolumn{5}{c|}{Aroma} & \multicolumn{5}{c}{Palate}\\
\cline{4-18}
\multicolumn{3}{c|}{} &\multicolumn{1}{c}{S}& \multicolumn{1}{c|}{Acc} & \multicolumn{1}{c}{P} & \multicolumn{1}{c}{R} &\multicolumn{1}{c|}{F1} &\multicolumn{1}{c}{S}& \multicolumn{1}{c|}{Acc} & \multicolumn{1}{c}{P} & \multicolumn{1}{c}{R} &\multicolumn{1}{c|}{F1} &\multicolumn{1}{c}{S}& \multicolumn{1}{c|}{Acc}& \multicolumn{1}{c}{P} & \multicolumn{1}{c}{R} &\multicolumn{1}{c}{F1}\\
\hline
\multicolumn{3}{c|}{RNP$^{*}$} &10.0&\multicolumn{1}{c|}{-}& 32.4 & 18.6 & 23.6  &10.0&\multicolumn{1}{c|}{-}&44.8 &32.4& 37.6 & 10.0 & \multicolumn{1}{c|}{-} & 24.6 & 23.5 & 24.0\\
\multicolumn{3}{c|}{INVRAT$^{*}$} & 10.0 & - & 42.6 &31.5 & 36.2 &10.0&- & 41.2 & 39.1 & 40.1&10.0&-&34.9&45.6&39.5\\
\multicolumn{3}{c|}{Inter-RAT$^{*}$} & 11.7 & \multicolumn{1}{c|}{-}&66.0 &46.5 &  {54.6} &11.7 & \multicolumn{1}{c|}{-} & 55.4 & 47.5&51.1&12.6&\multicolumn{1}{c|}{-}&34.6&48.2&40.2\\
\multicolumn{3}{c|}{FR} & 11.1 & 75.8&70.4 &42.0 & {52.6} &9.7 & 87.7 & 68.1 & 42.2& {52.1}&11.7&87.9&43.7&40.9& {42.3}\\

\multicolumn{3}{c|}{MCD-KL} &9.5&79.7&{94.2}&{48.4}&\underline{63.9}&9.9&87.5&\textbf{84.6}&\textbf{53.9}&\textbf{65.8}&9.4&87.3&{60.9}&{47.1}&\underline{53.1}
  \\
\multicolumn{3}{c|}{MCD-JS} &9.7&80.1&\textbf{95.7}&\textbf{50.2}&\textbf{65.9}&10.0&86.1&{79.8}&{51.0}&\underline{62.2}&10.9&85.6&\textbf{62.1}&\textbf{54.4}&\textbf{58.0}\\
  \hline
  \hline
  \multicolumn{3}{c|}{RNP$^{*}$} &20.0&\multicolumn{1}{c|}{-}& 39.4 & 44.9 & 42.0  &20.0&\multicolumn{1}{c|}{-}&37.5&51.9&43.5  & 20.0 & \multicolumn{1}{c|}{-}& 21.6 & 38.9 & 27.8\\
\multicolumn{3}{c|}{INVRAT$^{*}$} & 20.0 & - & 58.9 &67.2 & 62.8 &20.0&- & 29.3 & 52.1 & 37.5&20.0&-&24.0&55.2&33.5\\
\multicolumn{3}{c|}{Inter-RAT$^{*}$} & 21.7 & \multicolumn{1}{c|}{-}&62.0 &76.7 & 68.6 &20.4 & \multicolumn{1}{c|}{-} & 44.2 & 65.4&52.8&20.8&\multicolumn{1}{c|}{-}&26.3&59.1&36.4\\
\multicolumn{3}{c|}{FR} & 20.9 & 84.6&74.9&84.9 &  {79.6} &19.5 & 89.3 & 58.7 &73.3& {65.2}&20.2&88.2&36.6&59.4&45.3\\
\multicolumn{3}{c|}{MCD-KL} &20.0&85.5&\textbf{79.3}&\textbf{85.5}&\textbf{82.3}&19.3&88.4&\textbf{65.8}&\textbf{81.4}&\textbf{72.8}&19.6&{87.7}&{41.3}&{65.0}&\underline{50.5}
  \\
\multicolumn{3}{c|}{MCD-JS} &19.9&80.8&{77.7}&{83.4}&\underline{80.5}&
18.8&87.2&{60.5}&{73.1}&\underline{66.2}&20.2&{86.0}&\textbf{42.3}&\textbf{68.5}&\textbf{52.3}
  \\
  \hline
  \hline
    \multicolumn{3}{c|}{RNP$^{*}$} &30.0&\multicolumn{1}{c|}{-}& 24.2 & 41.2 & 30.5  &30.0&\multicolumn{1}{c|}{-}&27.1 &55.7& 36.4& 30.0 &\multicolumn{1}{c|}{-}& 15.4 & 42.2 & 22.6 \\
\multicolumn{3}{c|}{INVRAT$^{*}$} & 30.0 & - & 41.5 &74.8 & 53.4 &30.0&- & 22.8 & 65.1 & 33.8&30.0&-&20.9&{71.6}&32.3\\
\multicolumn{3}{c|}{Inter-RAT$^{*}$} & 30.5 & \multicolumn{1}{c|}{-}&48.1 &82.7 & 60.8 &29.4 & \multicolumn{1}{c|}{-} & 37.9 & 72.0&49.6&30.4&\multicolumn{1}{c|}{-}&21.8&{66.1}&{32.8}\\
\multicolumn{3}{c|}{FR} & 29.6 & 86.4&50.6 &81.4 &  {62.3}&30.8 & 88.1 & 37.4 & 75.0&49.9&30.1&87.0&24.5&58.8& {34.6}\\
\multicolumn{3}{c|}{MCD-KL}&29.7&86.7&{59.6}&\textbf{95.6}&\underline{73.4}&29.6&90.2&{46.1}&\textbf{87.5}&\underline{60.4}&29.4&{87.0}&\textbf{30.5}&\textbf{72.4}&\textbf{42.9}\\
\multicolumn{3}{c|}{MCD-JS} &29.0&89.6&\textbf{60.2}&{94.4}&\textbf{73.5}&28.7&86.2&\textbf{47.3}&{87.0}&\textbf{61.3}&27.6&{84.5}&{26.9}&{59.7}&\underline{37.1}
  \\
  \hline
\end{tabular}
}
   
    \label{tab:js}
\end{table*}

\subsection{Time efficiency}\label{app:time efficiency}

By avoiding many local optima, our MCD can converge much faster than MMI-based methods. 
Figure~\ref{fig:convergence speed} shows a comparison of convergence speed between our MCD and the latest MMI-based SOTA FR on \emph{Beer-Appearance} and \emph{Beer-Aroma} with $S\approx20$, where FR and MCD get the similar F1, and they use the same learning rate (0.0001) and batchsize (128).

\section{Proofs}
\subsection{Derivation of Equation~\ref{eqa:other bernoulli}\label{proof:other bernoulli}
}
We use $X_S$ as an example, and the others are nothing different.
\begin{equation}
   p(X_S=1)=\sum_{U\in \{0,1\}} p(X_S=1,U)=\sum_{U\in \{0,1\}} p(X_S=1|U)p(U)=0.9*0.5+0.1*0.5=0.5.
\end{equation}

\subsection{Derivation of Equation~\ref{eqa:correlated_aspects} and~\ref{eqa:taste2smell} 
}\label{proof:correlated_aspects}
In Figure~\ref{fig:causal}(a), we have $X_T\upmodels X_S|U$ and $X_T\upmodels Y_S|X_S$ (please refer to Appendix~\ref{app: example of probabilistic graph}). That is to say, 
\begin{equation}
    P(X_S|U,X_T)= P(X_S|U), \ P(Y_S|X_S,X_T)=P(Y_S|X_S).
\end{equation}

Then we can easily get Equation~\ref{eqa:correlated_aspects}:
\begin{equation}\label{eqa:proof correlated_aspects}
\begin{aligned}
    p(X_S=1|X_T=1)=&\sum_{U\in\{0,1\}}p(X_S=1,U|X_T=1)\\
    =&\sum_{U\in\{0,1\}}p(X_S=1|U,X_T=1)p(U|X_T=1)\\
    =&\sum_{U\in\{0,1\}}p(X_S=1|U)p(U|X_T=1).
\end{aligned}
\end{equation}
And  Equation~\ref{eqa:taste2smell} is similar.

\subsection{The relation between entropy and cross-entropy}\label{app:relation between entropy and cross-entropy}

It is a basic idea in information theory that the entropy of a distribution $P$ is upper bounded by the cross entropy of using $Q$ to approximate it.
For any two distribution $P$ and $Q$, we have 
\begin{equation}
\begin{aligned}
    H_c(P,Q)=H(P)+D_{KL}(P||Q)\geq H(P),
\end{aligned}
\end{equation}
where the subscript $c$ in $H_c(P,Q)$ stands for cross-entropy.

We know that we get the minimum cross entropy when $Q$ is the same as $P$, i.e., $D_{KL}(P||Q)=0$. Which means 
\begin{equation}\label{eqa:hch}
    \mathop{\min}H_c(P,Q)=H(P).
\end{equation}

\subsection{Conditional independence in a probabilistic graph}\label{app: example of probabilistic graph}

\begin{wrapfigure}[13]{R}{0.4\columnwidth}
    \includegraphics[width=0.4\columnwidth]{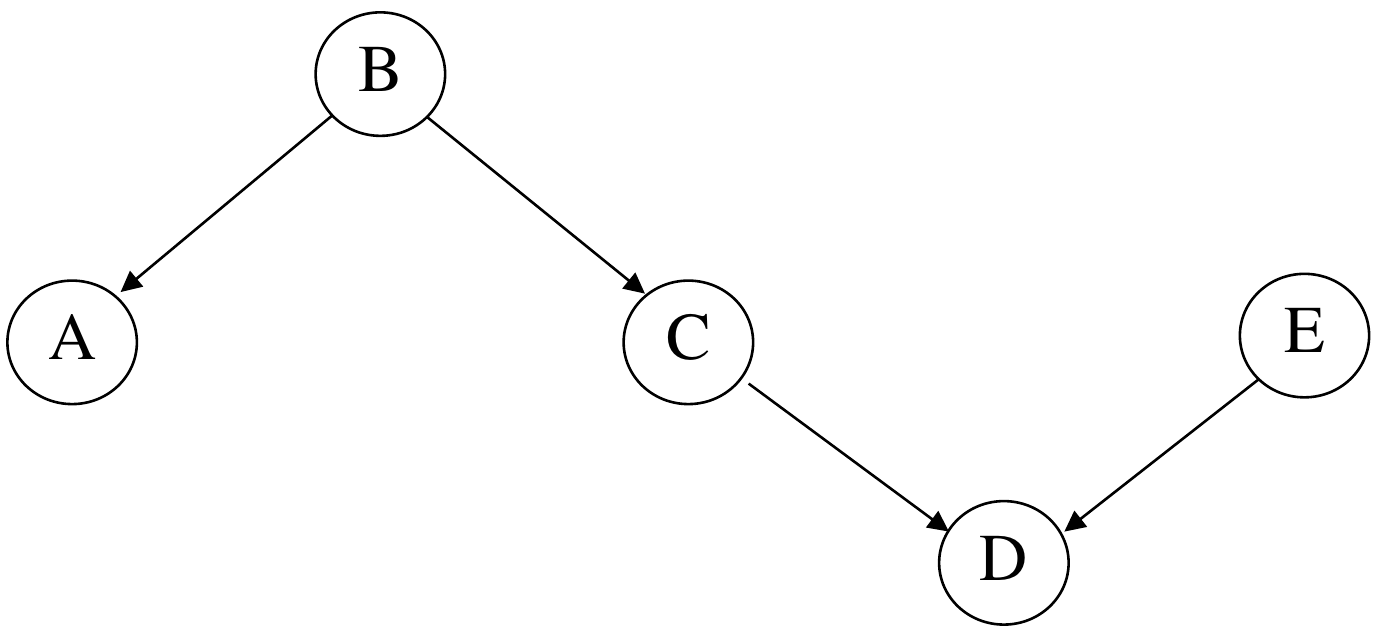}
  \caption{A probabilistic graph that contains a fork, a chain, and a collider.}
  \label{fig:probabilistic graph}
\end{wrapfigure}

In the probabilistic graph depicted in Figure~\ref{fig:probabilistic graph}, we have that $A\upmodels C|B$, $B\upmodels D|C$, and $C\upmodels E$ (but note that we do not have $C\upmodels E|D$). This property is fundamental in probabilistic graphical models. The proof is straightforward, and we illustrate it using $A\upmodels C|B$ as an example.

Based on the general principle of the chain rule, we can have 
\begin{equation}\label{eqa: general distribution}
\begin{aligned}
    P(A,B,C)=&P(C|A,B)P(A,B)\\
    =&P(C|A,B)P(A|B)P(B).
\end{aligned}
\end{equation}
Based on the graph structure in Figure~\ref{fig:probabilistic graph}, we have 
\begin{equation}\label{eqa:probabilistic graph distribution}
    P(A,B,C)=P(B)P(A|B)P(C|B)
\end{equation}
Combining Equation~\ref{eqa: general distribution} and~\ref{eqa:probabilistic graph distribution}, we get 
\begin{equation}
    P(C|B)=P(C|A,B),
\end{equation}
which means $A\upmodels C|B$.

If you are seeking a more intuitive understanding of blocked path, please refer to a concept called $``$Bayes ball$"$~\citep{Jordan-2003}.

\subsection{Proof of Lemma~\ref{lem:d-sep to causal}}\label{app:proof of d-sep to causal}
To prove this, we employ a proof by contradiction. We initially assume that $X_C$ is a variable in $X_R$ and $X_C \notin X_Z$. Given that $X_C\in X_R$, we deduce that $X_C$ exerts a direct causal influence on $Y$, i.e., there exists a path $X_C\xrightarrow{}Y$:

\begin{equation*}
    X_C\in X_R \Longrightarrow X_C\xrightarrow{}Y.
\end{equation*}

Furthermore, since $X_C \notin X_Z$, we ascertain that $X_C \in X_{-Z}$. Consequently, we understand that $X_{-Z}$ and $Y$ are not d-separated by $X_Z$ due to an unblocked path $X_C\xrightarrow{}Y$.

With this, we complete the proof of Lemma~\ref{lem:d-sep to causal}.

\subsection{Failure cases of Assumption~\ref{assumption:no causal from y to x}}\label{app: failure of ass1}

 As far as we know, most of the real-world datasets are built in a collecting-annotating form. In such a form, $Y$ is given according to $X$, and the annotators won't edit $X$ after giving $Y$. So, Assumption~\ref{assumption:no causal from y to x} holds.
But there are also some cases that might break Assumption~\ref{assumption:no causal from y to x}. One is the synthetic data like ColorMinist. In ColorMinist, a human first annotates an image, and then edits the image again according to the assigned label. Another scenario is the collection of time series data, where annotators label the data based on existing information and then adjust the data collection method according to the previous labels. This creates a cyclic causal graph. However, in the literature of causal inference, most researchers only consider acyclic graphs. 
We note that in cases where Assumption~\ref{assumption:no causal from y to x} doesn't hold, we still have Lemma~\ref{lem:d-sep to causal}, i.e., D-separation severs as a sufficient condition for selecting causal rationales. When Assumption~\ref{assumption:no causal from y to x} holds, it becomes a necessary and sufficient condition.

\subsection{Proof of Lemma~\ref{lem:causal to d-sep}}\label{app: proof of causal to d-sep}
To prove it, we employ a proof by contradiction. We first assume a variable $X_C \in X_{-Z}$, and $X_C$ is associated with $Y$ conditioned on $X_Z$. To achieve the association, there must be a path in either of the following two forms. The first form is  
\begin{equation}
   X_C \cdots o \xrightarrow{} Y, \quad \ s.t.\  o \notin X_Z,  
\end{equation}
where $``\cdots"$ denotes some arbitrary  arrows and nodes, and $o$ is a intermediate node. $o \notin X_Z$ is from that if $o \in X_Z$, the path will be blocked by $X_Z$.

Since $o$ is a direct cause of $Y$, we have $o \in X_R$. Since $X_R\subset X_Z$, but we have $o \notin X_Z$, so this form of paths do not exist. 

The second form is 
\begin{equation}
   X_C \cdots \xrightarrow{} o \xleftarrow{}o_1\xleftarrow{}\cdots o_n\xleftarrow{} Y, \quad \ s.t.\  o \in X_Z,  
\end{equation}
where $o_1\cdots o_n$ are some nodes connected by left arrows, we do not discuss these nodes since discussing $o$ is enough for our proof.

This path is unblocked through a collider. Note that the way to unblock a collider path is to condition on it, so we need to have $o \in X_Z$. However, in this case, $Y$ has a causal effect on $o$, which breaks Assumption~\ref{assumption:no causal from y to x}.  So, this form of paths do not exist as well. 

As a result, there is no variable in $X_{-Z}$ can be associated with $Y$ conditioned on $X_Z$. The proof of Lemma~\ref{lem:causal to d-sep} is completed.

\end{document}